\documentclass[10pt,twocolumn,letterpaper]{article}

\usepackage{titling}
\usepackage{cvpr}
\usepackage{times}
\usepackage{epsfig}
\usepackage{graphicx}
\usepackage{amsmath}
\usepackage{amssymb}
\usepackage{lib}
\usepackage{mathtools}
\usepackage{multirow}
\usepackage{booktabs}
\usepackage{subfig}
\usepackage{balance}
\usepackage[font=small,labelfont=bf]{caption}
\usepackage[urlcolor=red,pagebackref=true,breaklinks=true,letterpaper=true,colorlinks,bookmarks=false]{hyperref}

\cvprfinalcopy 

\newcommand\blfootnote[1]{%
  \begingroup
  \renewcommand\thefootnote{}\footnote{#1}%
  \addtocounter{footnote}{-1}%
  \endgroup
}

\begin{document}


\title{Multi-view Supervision for Single-view Reconstruction\\ via Differentiable Ray Consistency
}
\author{Shubham Tulsiani, Tinghui Zhou, Alexei A. Efros, Jitendra Malik  \\
University of California, Berkeley\\
{\tt\small\{shubhtuls,tinghuiz,efros,malik\}@eecs.berkeley.edu}
}
\maketitle

\begin{abstract}
We study the notion of consistency between a 3D shape and a 2D observation and propose a differentiable formulation which allows computing gradients of the 3D shape given an observation from an arbitrary view. We do so by reformulating view consistency using a differentiable ray consistency (DRC) term. We show that this formulation can be incorporated in a learning framework to leverage different types of multi-view observations \eg foreground masks, depth, color images, semantics \etc as supervision for learning single-view 3D prediction. We present empirical analysis of our technique in a controlled setting. We also show that this approach allows us to improve over existing techniques for single-view reconstruction of objects from the PASCAL VOC dataset.
\end{abstract}

\blfootnote{Project website with code: \url{https://shubhtuls.github.io/drc/}}

\vspace{-4mm}
\section{Introduction}
\vspace{-1mm}

When is a solid 3D shape consistent with a 2D image?  If it is not, how do we change it to make it more so? One way this problem has been traditionally addressed is by space carving~\cite{carving}. Rays are projected out from pixels into the 3D space and each ray that is known not to intersect the object removes the volume in its path, thereby making the carved-out shape consistent with the observed image.

But what if we want to extend this notion of consistency to the differential setting?  That is, instead of deleting chunks of volume all at once, we would like to compute incremental changes to the 3D shape that make it more consistent with the 2D image.  In this paper, we present a differentiable ray consistency formulation that allows computing the gradient of a predicted 3D shape of an object, given an observation (depth image, foreground mask, color image \etc.) from an arbitrary view.

The question of finding a differential formulation for ray consistency is mathematically interesting in and of itself.  Luckily, it is also extremely useful as it allows us to connect the concepts in 3D geometry with the latest developments in machine learning.  While classic 3D reconstruction methods require large number of 2D views of the same physical 3D object, learning-based methods are able to take advantage of their past experience and thus only require a small number of views for each physical object being trained.  
Finally, when the system is done learning, it is able to give an estimate of the 3D shape of a novel object from only a \emph{single} image, something that classic methods are incapable of doing.  The differentiability of our consistency formulation is what allows its use in a learning framework, such as a neural network. Every new piece of evidence gives gradients for the predicted shape, which, in turn, yields incremental updates for the underlying prediction model.  Since this prediction model is shared across object instances, it is able to find and learn from the commonalities across different 3D shapes, requiring only sparse per-instance supervision.

\begin{figure*}[ht!]
\centering
\includegraphics[width=1.0\textwidth]{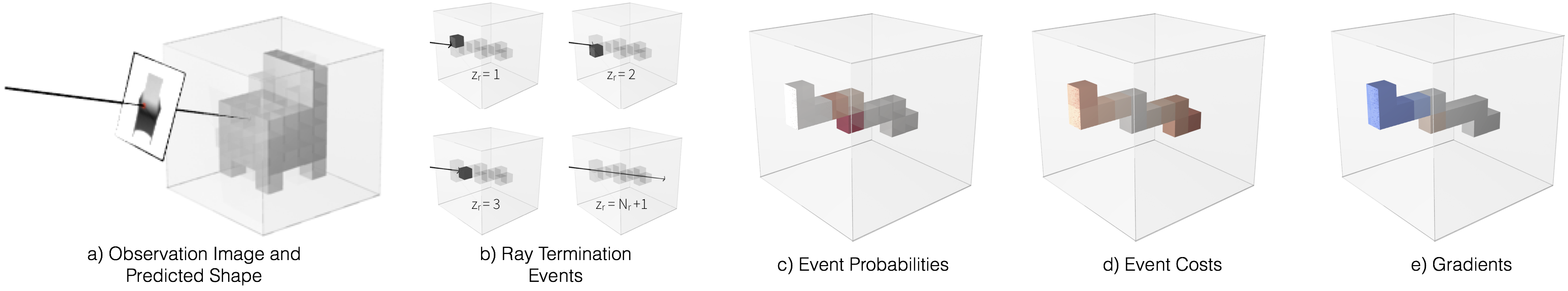}

\vspace{-2mm}
\caption{\small Visualization of various aspects of our Differentiable Ray Consistency formulation. a) Predicted 3D shape represented as probabilistic occupancies and the observation image where we consider consistency between the predicted shape and the ray corresponding to the highlighted pixel. b) Ray termination events (\secref{raytrace}) -- the random variable $z_r = i$ corresponds to the event where the ray terminates at the $i^{th}$ voxel on its path, $z_r = N_r+1$ represents the scenario where the ray escapes the grid. c) Depiction of event probabilities (\secref{raytrace}) where red indicates a high probability of the ray terminating at the corresponding voxel. d) Given the ray observation, we define event costs (\secref{eventcosts}). In the example shown, the costs are low (white color) for events where ray terminates in voxels near the observed termination point and high (red color) otherwise. e) The ray consistency loss (\secref{consloss}) is defined as the expected event cost and our formulation allows us to obtain gradients for occupancies (red indicates that loss decreases if occupancy value increases, blue indicates the opposite). While in this example we consider a depth observation, our formulation allows incorporating diverse kinds of observations by defining the corresponding event cost function as discussed in \secref{eventcosts} and \secref{labelFormulation}. Best viewed in color.}
\vspace{-3mm}
\figlabel{formulation}
\end{figure*}

\vspace{-1mm}
\section{Related Work}
\vspace{-1mm}
\seclabel{related}

\noindent \textbf{Object Reconstruction from Image-based Annotations.} Blanz and Vetter ~\cite{blanz1999morphable} demonstrated the use of a morphable model to capture 3D shapes.  Cashman and Fitzgibbon ~\cite{cashman2013shape} learned these models  for complex categories like dolphins using object silhouettes and keypoint annotations for training and inference. Tulsiani \etal ~\cite{pamishapeTulsianiKCM15} extended similar ideas to more general categories and leveraged recognition systems~\cite{girshick2014rich,hariharan2015hypercolumns,vpsKpsTulsianiM15} to automate test-time inference. Wu \etal~\cite{wu2016single}, using similar annotations, learned a system to predict sparse 3D by inferring parameters of a shape skeleton. However, since the use of such low-dimensional models restricts expressivity, Vicente \etal~\cite{vicente2014reconstructing} proposed a non-parametric method by leveraging  surrogate instances -- but at the cost of requiring annotations at test time. We leverage similar training data but using a CNN-based voxel prediction framework allows test time inference without manual annotations and allows handling large shape variations.

\vspace{2mm}
\noindent \textbf{Object Reconstruction from 3D Supervision.}
The advent of deep learning along with availability of large-scale synthetic training data has resulted in applications for object reconstruction. 
Choy \etal ~\cite{choy20163d} learned a CNN to predict a voxel representation using a single (or multiple) input image(s). Girdhar \etal ~\cite{Girdhar16b} also presented similar results for single-view object reconstruction, while also demonstrating some results on real images by using realistic rendering techniques~\cite{su2015render} for generating training data. A crucial assumption in the procedure of training these models, however, is that full 3D supervision is available. As a result, these methods primarily train using synthetically rendered data where the underlying 3D shape is available.

While the progress demonstrated by these methods is encouraging and supports the claim for using CNN based learning techniques for reconstruction, the requirement of explicit 3D supervision for training is potentially restrictive. We relax this assumption and show that alternate sources of supervision can be leveraged. It allows us to go beyond reconstructing objects in a synthetic setting, to extend to real datasets which do not have 3D supervision.

\vspace{2mm}
\noindent \textbf{Multi-view Instance Reconstruction.}
Perhaps most closely related to our work in terms of the proposed formulation is the line of work in geometry-based techniques for reconstructing a single instance given multiple views. Visual hull~\cite{laurentini1994visual} formalizes the notion of consistency between a 3D shape and observed object masks. Techniques based on this concept~\cite{broadhurst2001probabilistic,matusik2000image} can obtain reconstructions of objects by space carving using multiple available views. It is also possible, by jointly modeling appearance and occupancy, to recover 3D structure of objects/scenes from multiple images via ray-potential based optimization~\cite{de1999roxels,liu2010ray} or inference in a generative model~\cite{gargallo2007occupancy}. Ulusoy~\etal~\cite{ulusoy2015towards} propose a probabilistic framework where marginal distributions can be efficiently computed. More detailed reconstructions can be obtained by incorporating additional signals \eg depth or semantics~\cite{kundu2014joint,savinov2016semantic,savinov2015discrete}.

The main goal in these prior works is to reconstruct a specific scene/object from multiple observations and they typically infer a discrete assignment of variables such that it is maximally consistent with the available views.
Our insight is that similar cost functions which measure consistency, adapted to treat variables as continuous probabilities, can be used in a learning framework to obtain gradients for the current prediction. Crucially, the multi-view reconstruction approaches typically solve a (large) optimization to reconstruct a particular scene/object instance and require a large number of views. In contrast, we only need to perform a single gradient computation to obtain a learning signal for the CNN and can even work with sparse set of views (possibly even just one view) per instance.

\vspace{2mm}
\noindent \textbf{Multi-view Supervision for Single-view Depth Prediction.}
While single-view depth prediction had been dominated by approaches with direct supervision~\cite{eigen2015predicting}, recent approaches based on multi-view supervision have shown promise in achieving similar (and sometimes even better) performance. Garg \etal~\cite{garg2016unsupervised} and Godard \etal~\cite{godard2016unsupervised} used stereo images to learn a single image depth prediction system by minimizing the inconsistency as measured by pixel-wise reprojection error. Zhou \etal~\cite{zhou2017unsupervised} further relax the constraint of having calibrated stereo images, and learn a single-view depth model from monocular videos. The motivation of these multi-view supervised depth prediction approaches is similar to ours, but we aim for 3D instead of 2.5D predictions and address the related technical challenges in this work.

\vspace{-1mm}
\section{Formulation}
\vspace{-1mm}
\seclabel{formulation}
In this section, we formulate a differentiable `view consistency' loss function which measures the inconsistency between a (predicted) 3D shape and a corresponding observation image. We first formally define our problem setup by instantiating the representation of the 3D shape and the observation image with which the consistency is measured.

\vspace{2mm}
\noindent \textbf{Shape Representation.}
Our 3D shape representation is parametrized as occupancy probabilities of cells in a dicretized 3D voxel grid, denoted by the variable $x$. We use the convention that $x_i$ represents the probability of the $i^{th}$ voxel being empty (we use the term `occupancy probability' for simplicity even though it is a misnomer as the variable $x$ is actually `emptiness probability'). Note that the choice of discretization of the 3D space into voxels need not be a uniform grid -- the only assumption we make is that it is possible to trace rays across the voxel grid and compute intersections with cell boundaries.

\vspace{2mm}
\noindent \textbf{Observation.} We aim for the shape to be consistent with some available observation $O$. This `observation' can take various forms \eg  a depth image, or an object foreground mask -- these are treated similarly in our framework. Concretely, we have a observation-camera pair $(O, C)$ where the `observation' $O$ is from a view defined by camera $C$.

\vspace{1mm}
Our view consistency loss, using the notations mentioned above, is of the form $L(x;(O,C))$. In \secref{rbc}, we reduce the notion of consistency between the 3D shape and an observation image to consistency between the 3D shape and a ray with associated observations. We then proceed to present a differentiable formulation for ray consistency, the various aspects of which are visualized in \figref{formulation}. In \secref{raytrace}, we examine the case of a ray travelling though a probabilistically occupied grid and in \secref{eventcosts}, we instantiate costs for each probabilistic ray-termination event. We then combine these to define the consistency cost function in \secref{consloss}. While we initially only consider the case of the shape being represented by voxel occupancies $x$, we show in \secref{labelFormulation} that it can be extended to incorporate optional per-voxel predictions $p$. This generalization allows us to incorporate other kinds of observation \eg color images, pixel-wise semantics \etc.
The generalized consistency loss function is then of the form $L(x,[p]; (O, C) )$ where $[p]$ denotes an optional argument.

\subsection{View Consistency as Ray Consistency}
\seclabel{rbc}
Every pixel in the observation image $O$ corresponds to a ray with a recorded observation (depth/color/foreground label/semantic label).  Assuming known camera intrinsic parameters ($f_u,f_v,u_0,v_0$), the image pixel $(u,v)$ corresponds to a ray $r$ originating from the camera centre travelling in direction $(\frac{u-u_0}{f_u}, \frac{v-v_0}{f_v},1)$ in the camera coordinate frame. Given the camera extrinsics, the origin  and direction of the ray $r$ can also be inferred in the world frame.

Therefore, the available observation-camera pair $(O, C)$ is equivalently a collection of arbitrary rays $\mathcal{R}$ where each $r \in \mathcal{R}$ has a known origin point, direction and an associated observation $o_r$ \eg depth images indicate the distance travelled before hitting a surface, foreground masks inform  whether the ray hit the object, semantic labels correspond to observing category of the object the ray terminates in.

We can therefore formulate the view consistency loss $L(x; (O, C))$ using per-ray based consistency terms $L_r(x)$. Here, $L_r(x)$ captures if the inferred 3D model $x$ correctly explains the observations associated with the specific ray $r$. Our view consistency loss is then just the sum of the consistency terms across the rays:
\begin{gather}
\eqlabel{rbc}
L(x; (O, C)) \equiv \underset{r \in \mathcal{R}}{\sum} L_r(x)
\end{gather}
Our task for formulating the view consistency loss is simplified to defining a differentiable ray consistency loss $L_r(x)$.

\subsection{Ray-tracing in a Probabilistic Occupancy Grid}
\seclabel{raytrace}
With the goal of defining the consistency cost $L_r(x)$, we examine the ray $r$ as it travels across the voxel grid with occupancy probabilities $x$. The motivation is that a probabilistic occupancy model (instantiated by the shape parameters $x$) induces a distribution of \textit{events} that can occur to ray $r$ and we can define $L_r(x)$ by seeing the incompatibility of these events with available observations $o_r$.

\vspace{2mm}
\noindent \textbf{Ray Termination Events.}
Since we know the origin and direction for the ray $r$, we can trace it through the voxel grid - let us assume it passes though $N_r$ voxels. The \textit{events} associated with this ray correspond to it either terminating at one of these $N_r$ voxels or passing through. We use a random variable $z_r$ to  correspond to the voxel in which the ray (probabilistically) terminates - with $z_r = N_r + 1$ to represent the case where the ray does not terminate. These events are shown in \figref{formulation}.

\vspace{2mm}
\noindent \textbf{Event Probabilities.}
Given the occupancy probabilities $x$, we want to infer the probability $p(z_r = i)$. The event $z_r = i$ occurs iff the previous voxels in the path are all unoccupied and the $i^{th}$ voxel is occupied. Assuming an independent distribution of occupancies where the prediction $x^{r}_i$ correspnds to the probability of the $i^{th}$ voxel on the path of the ray $r$ as being \textit{empty}, we can compute the probability distribution for $z_r$.
\begin{gather}
\eqlabel{pzr}
p(z_r = i) = \begin{dcases}
    (1 - x^r_i) \prod_{j=1}^{i-1} x^r_j, & \text{if } i \leq N_r\\
     \quad \prod_{j=1}^{N_r} x^r_j,  & \text{if } i = N_r+1
\end{dcases}
\end{gather}

\subsection{Event Cost Functions}
\seclabel{eventcosts}

Note that each event $(z_r = i)$, induces a prediction \eg if $z_r = i$, we can geometrically compute the distance $d^r_i$ the ray travels before terminating.
We can define a cost function between the induced prediction under the event $(z_r = i)$ and the available associated observations for ray $o_r$. We denote this cost function as $\psi_r(i)$ and it assigns a cost to event $(z_r = i)$ based on whether it induces predictions inconsistent with $o_r$. We now show some examples of event cost functions that can incorporate diverse observations $o_r$ and used in various scenarios.

\vspace{2mm}
\noindent \textbf{Object Reconstruction from Depth Observations.}
In this scenario, the available observation $o_r$ corresponds to the observed distance the ray travels $d_r$.
We use a simple distance measure between observed distance and event-induced distance to define $\psi_r(i)$.
\begin{gather}
\eqlabel{psidepth}
\psi^{depth}_r(i) = | d^r_i - d_r |
\end{gather}

\vspace{2mm}
\noindent \textbf{Object Reconstruction from Foreground Masks.}
We examine the case where we only know the object masks from various views. In this scenario, let $s_r \in \{0,1\}$ denote the known information regarding each ray - $s_r=0$ implies the ray $r$ intersects the object \ie corresponds to an image pixel within the mask, $s_r=1$ indicates otherwise. We can capture this by defining the corresponding cost terms.
\begin{gather}
\psi^{mask}_r(i) = 
\begin{dcases}
    s_r, & \text{if } i \leq N_r\\
    1-s_r,  & \text{if } i = N_r+1
\end{dcases}
\end{gather}
We note that some concurrent approaches~\cite{rezende2016unsupervised,yan2016perspective} have also been proposed to specifically address the case of learning object reconstruction from foreground masks. These approaches, either though a learned~\cite{rezende2016unsupervised} or fixed~\cite{yan2016perspective} reprojection function, minimize the discrepancy between the observed mask and the reprojected predictions. We show in the appendix
that our ray consistency based approach effectively minimizes a similar loss using a geometrically derived re-projection function, while also allowing us to handle more general observations.

\subsection{Ray-Consistency Loss}
\seclabel{consloss}
We have examined the case of a ray traversing through the probabilistically occupied voxel grid and defined possible ray-termination events occurring with probability distribution specified by $p(z_r)$. For each of these events, we incur a corresponding cost $\psi_r(i)$ which penalizes inconsistency between the event-induced predictions and available observations $o_r$. The per-ray consistency loss function $L_r(x)$ is simply the expected cost incurred.
\begin{gather}
\eqlabel{Lrdef1}
L_r(x) = \mathbb{E}_{z_r} [\psi_r(z_r)] \\
\eqlabel{Lrdef2}
L_r(x) = \sum_{i=1}^{N_r+1} \psi_r(i)~p(z_r = i) 
\end{gather}
Recall that the event probabilities $p(z_r=i)$ were defined in terms of the voxel occupancies $x$ predicted by the CNN (\eqref{pzr}). Using this, we can compute the derivatives of the loss function $L_r(x)$ w.r.t the CNN predictions (see Appendix for derivation).
\begin{gather}
\eqlabel{grad_x}
\frac{\partial~L_r(x)}{\partial~x^r_k} = \sum_{i=k}^{N_r} ~ (\psi_r(i+1) - \psi_r(i)) \prod_{1\leq j \leq i, j \neq k} x^r_j
\end{gather}
The ray-consisteny loss $L_r(x)$ completes our formulation of view consistency loss as the overall loss is defined in terms of $L_r(x)$ as in \eqref{rbc}. The gradients derived from the view consistency loss simply try to adjust the voxel occupancy predictions $x$, such that events which are inconsistent with the observations occur with lower probabilities.

\subsection{Incorporating Additional Labels}
\seclabel{labelFormulation}
We have developed a view consistency formulation for the setting where the shape representation is described as occupancy probabilities $x$. In the scenario where alternate per-pixel observations (\eg semantics or color) are available, we can modify consistency formulation to account for per-voxel predictions $p$ in the 3D representation. In this scenario, the observation $o_r$ associated with the ray $r$ includes the corresponding pixel label and similarly, the induced prediction under event $(z_r = i)$  includes the auxiliary prediction for the $i^{th}$ voxel on the ray's path -- $p^r_i$.

To incorporate consistency between these, we can extend $L_r(x)$ to $L_r(x,[p])$ by using a generalized event-cost term $\psi_r(i,[p^r_i])$ in \eqref{Lrdef1} and \eqref{Lrdef2}. Examples of the generalized cost term for two scenarios are presented in \eqref{psisem} and \eqref{psicolor}. The gradients for occupancy predictions $x_i^r$ are as previously defined in \eqref{grad_x}, but using the generalized cost term $\psi_r(i,[p^r_i])$ instead. The additional per-voxel  predictions can also be trained using the derivatives below.
\begin{gather}
\eqlabel{grad_p}
\frac{\partial~L_r(x,[p])}{\partial~p^i_r} = p(z_r=i) \frac{\partial~\psi_r(i, [p^i_r])}{\partial~p^i_r}
\end{gather}

Note that we can define any event cost function $\psi(i, [p^r_i])$ as long as it is differentiable w.r.t $p^r_i$. We can interpret \eqref{grad_p} as the  additional per-voxel predictions $p$ being updated to match the observed pixel-wise labels, with the gradient being weighted by the probability of the corresponding event.

\vspace{2mm}
\noindent \textbf{Scene Reconstruction from Depth and Semantics.}
In this setting, the observations associated with each ray correspond to an observed depth $d_r$ as well as semantic class labels $c_r$. The event-induced prediction, if $z_r = i$, corresponds to depth $d^r_i$ and class distribution $p^r_i$ and we can define an event cost penalizing the discrepancy in disparity (since absolute depth can have a large variation) and the negative log likelihood of the observed class.
\begin{gather}
\eqlabel{psisem}
\psi^{sem}_r(i, p^r_i) =  | \frac{1}{d^r_i} -  \frac{1}{d_r} | - log(p^r_i(c_r))
\end{gather}

\vspace{2mm}
\noindent \textbf{Object Reconstruction from Color Images.}
In this scenario, the observations $c_r$ associated with each ray corresponds to the RGB color values for the corresponding pixel. Assuming additional per voxel color prediction $p$, the event-induced prediction, if $z_r = i$, yields the color at the corresponding voxel \ie $p^r_i$. We can define an event cost penalizing the squared error.
\begin{gather}
\eqlabel{psicolor}
\psi^{color}_r(i, p^r_i) =  \frac{1}{2} \| p^r_i - c_r \|^2
\end{gather}
In addition to defining the event cost functions, we also need to instantiate the induced observations for the event of ray escaping. We define $d^r_{N_r+1}$ in \eqref{psidepth} and \eqref{psisem} to be a fixed large value, and $p^r_{N_r+1}$ in \eqref{psisem} and \eqref{psicolor} to be uniform distribution and white color respectively. We discuss this further in the appendix.

\vspace{-1mm}
\section{Learning Single-view Reconstruction}
\vspace{-1mm}
\seclabel{learning}
We aim to learn a function $f$ — modeled as a parameterized CNN $f_{\theta},$ which given a single image $I$ corresponding to a novel object, predicts its shape as a voxel occupancy grid. A straightforward learning-based approach would require a training dataset $\{(I_i, \bar{x}_i)\}$ where the target voxel representation $\bar{x}_i$ is known for each training image $I_i$. However, we are interested in a scenario where the ground-truth 3D models $\{\bar{x}_i\}$ are not available for training $f_{\theta}$ directly, as is often the case for real-world objects/scenes. While collecting the ground-truth 3D is not feasible, it is relatively easy to obtain 2D or 2.5D observations (e.g. depth maps) of the underlying 3D model from other viewpoints. In this scenario we can leverage the `view consistency' loss function described in \secref{formulation} to train  $f_{\theta}$ .

\vspace{2mm}
\noindent \textbf{Training Data.}
As our training data, corresponding to each training (RGB) image $I_i$ in the training set, we also have access to one or more additional observations of the same instance from other views. The observations, as described in \secref{formulation}, can be of varying forms. Concretely, corresponding to image $I_i$, we have \emph{one or more} observation-camera pairs $\{O^i_k, C^i_k\}$ where the `observation' $O^i_k$ is from a view defined by camera $C^i_k$. Note that these observations are required only for training; at test time, the learned CNN $f_{\theta}$ predicts a 3D shape from only a single 2D image.

\vspace{2mm}
\noindent \textbf{Predicted 3D Representation.} The output of our single-view 3D prediction CNN is $f_{\theta} (I) \equiv (x,[p])$ where $x$ denotes voxel occupancy probabilities and $[p]$ indicates optional per-voxel predictions (used if corresponding training observations \eg color, semantics are leveraged).

\vspace{2mm}
To learn the parameters $\theta$ of the single-view 3D prediction CNN, for each training image $I_i$ we train the CNN to minimize the inconsistency between the prediction $f_{\theta} (I_i)$ and the one or more observation(s) $\{(O^i_k, C^i_k)\}$ corresponding to $I_i$. This optimization is the same as minimizing the (differentiable) loss function $\underset{i}{\sum}~
\underset{k}{\sum}~
L(f_{\theta} (I_i); (O^i_k, C^i_k))$ \ie the sum of view consistency losses (\eqref{rbc}) for observations across the training set.
To allow for faster training, instead of using all rays as defined in \eqref{rbc}, we randomly sample a few rays (about 1000) per view every SGD iteration.

\vspace{-1mm}
\section{Experiments}
\vspace{-1mm}
\seclabel{experiments}
We consider various scenarios where we can learn single-view reconstruction using our differentiable ray consistency (DRC) formulation. First, we examine the ShapeNet dataset where we use synthetically generated images and corresponding multi-view observations to study our framework. We then demonstrate applications on the PASCAL VOC dataset where we train a single-view 3D prediction system using only one observation per training instance. We then explore the application of our framework for scene reconstruction using short driving sequences as supervision. Finally, we show qualitative results for using multiple color image observations as supervision for single-view reconstruction.

\subsection{Empirical Analysis on ShapeNet}
\begin{figure*}[t!]
\centering
\includegraphics[width=1.0\textwidth]{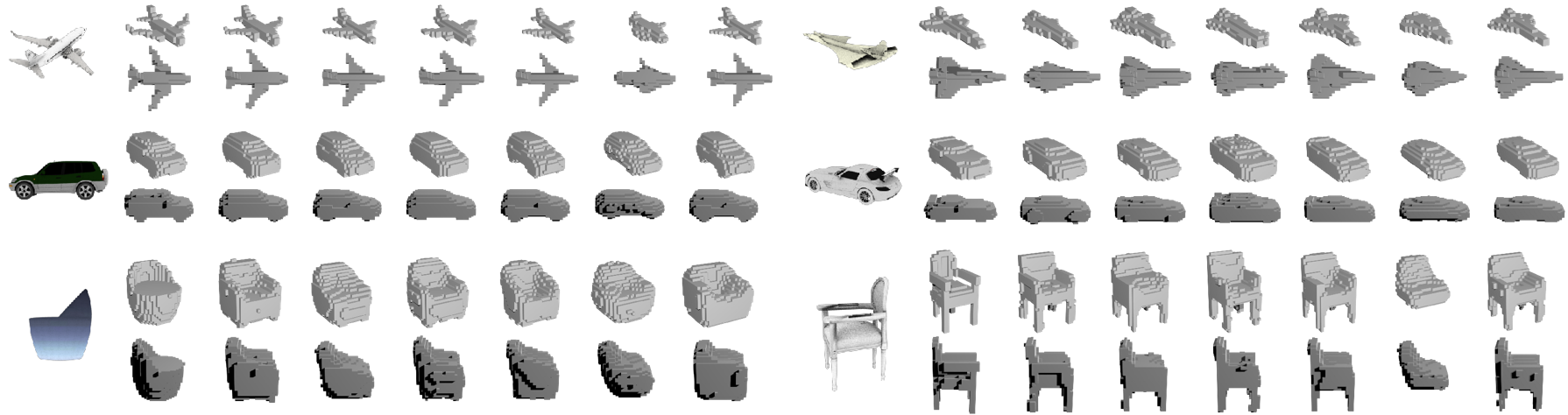}

\vspace{-2mm}
\caption{\small Reconstructions on the ShapeNet dataset visualized using two representative views. Left to Right : Input, Ground-truth, 3D Training, Ours (Mask), Fusion (Depth), DRC (Depth), Fusion (Noisy Depth), DRC (Noisy Depth).}
\figlabel{shapenetFig}
\vspace{-2mm}
\end{figure*}

We study the framework presented and demonstrate its applicability with different types of multi-view observations and also analyze the susceptibility to noise in the learning signal. We perform experiments in a controlled setting using synthetically rendered data where the ground-truth 3D information is available for benchmarking.

\begin{figure}
 \subfloat[Number of training views]{\includegraphics[width=0.49\linewidth]{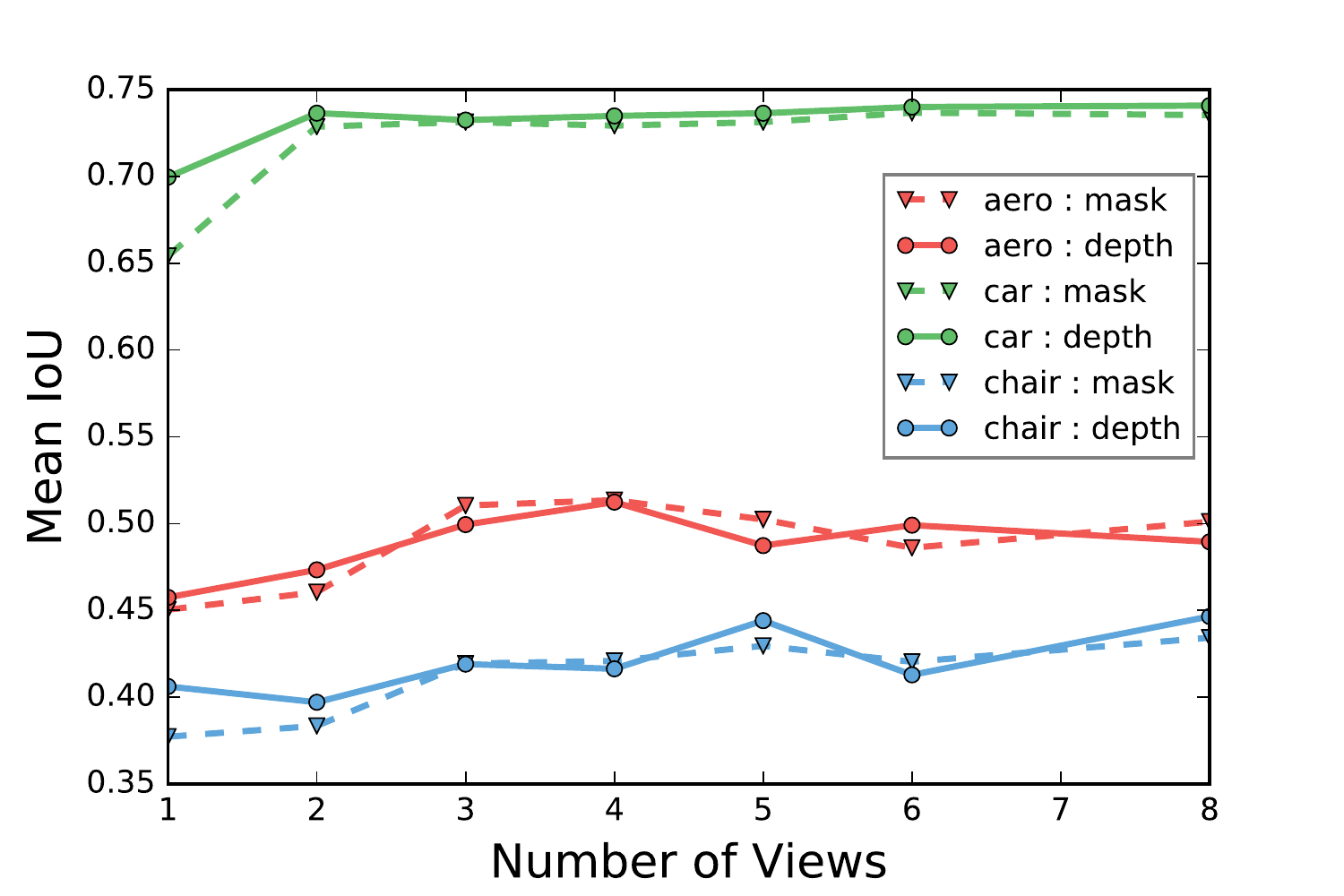}} \hfill
 \subfloat[Amount of noise]{\includegraphics[width=0.49\linewidth]{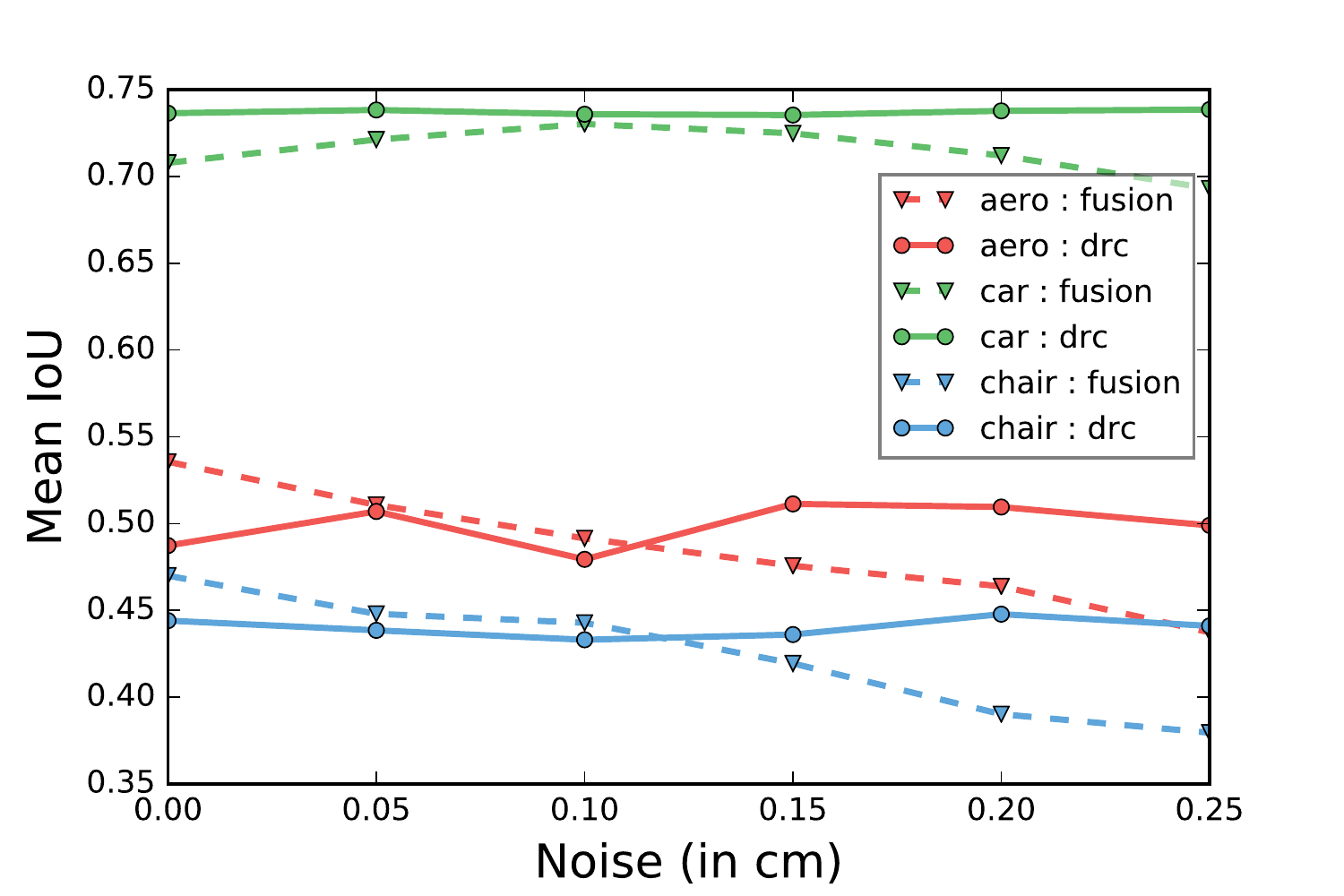}}

\vspace{-2mm}
\caption{\small Analysis of the per-category reconstruction performance. a) As we increase the number of views available per instance for training, the performance initially increases and saturates after few available views. b) As the amount of noise in depth observations used for training increases, the performance of our approach remains relatively consistent.}
\vspace{-2mm}
\figlabel{snetAnalysis}
\end{figure}

\vspace{2mm}
\noindent \textbf{Setup.}
The ShapeNet dataset~\cite{shapenet2015} has a collection of textured CAD models and we examine 3 representative categories with large sets of available models : airplanes, cars, and chairs . We create random train/val/test splits and use rendered images with randomly sampled views as input to the single-view 3D prediction CNNs.

Our CNN model is a simple encoder-decoder which predicts occupancies in a voxel grid from the input RGB image (see appendix for details). To perform control experiments, we vary the sources of information available (and correspondingly, different loss functions) for training the CNN. The various control settings are briefly described below (and explained in detail in the appendix) :

\vspace{1mm}
\noindent \textit{Ground-truth 3D.} We assume that the ground-truth 3D model is available and use a simple cross-entropy loss for training. This provides an upper bound for the performance of a multi-view consistency method.
    
\vspace{1mm}
\noindent \textit{DRC (Mask/Depth).} In this scenario, we assume that (possibly noisy) depth images (or object masks) from 5 random views are available for each training CAD model and minimize the view consistency loss.

\vspace{1mm}
\noindent \textit{Depth Fusion.} As an alternate way of using multi-view information, we preprocess the 5 available depth images per CAD model to compute a pseudo-ground-truth 3D model.
We then train the CNN with a cross-entropy loss, restricted to voxels where the views provided any information. Note that unlike our method, this is applicable only if depth images are available
and is more susceptible to noise in observations. See appendix for further details and discussion.

\begin{figure*}[t!]
\hspace{2mm}
\includegraphics[width=0.95\textwidth]{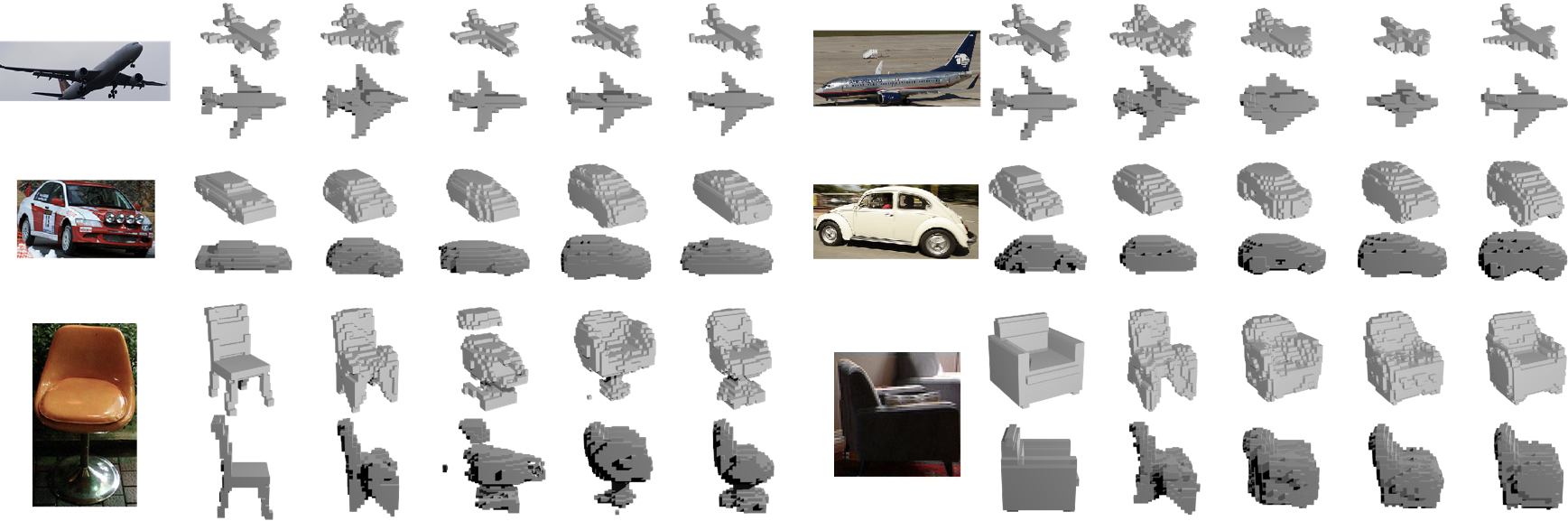}

\vspace{-2mm}
\caption{\small PASCAL VOC reconstructions visualized using two representative views. Left to Right : Input, Ground-truth (as annotated in PASCAL 3D), Deformable Models~\cite{pamishapeTulsianiKCM15}, DRC (Pascal), Shapenet 3D, DRC (Joint).}
\vspace{-3mm}
\figlabel{pascalFig}
\end{figure*}

\vspace{2mm}
\noindent \textbf{Evaluation Metric.} We use the mean intersection over union (IoU) between the ground-truth 3D occupancies and the predicted 3D occupancies. Since different losses lead to the learned models being calibrated differently, we report mean IoU at the optimal discretization threshold for each method (the threshold is searched at a category level).

\renewcommand{\arraystretch}{1.2}
\setlength{\tabcolsep}{4pt}
\begin{table}
\centering
\footnotesize
\resizebox{1.0\linewidth}{!}{
\begin{tabular}{lccccccc}
\toprule
Training Data & \multicolumn{1}{c}{3D} & \multicolumn{2}{c}{{Mask}} & \multicolumn{2}{c}{{Depth}} &  \multicolumn{2}{c}{{Depth (Noisy)}}
\\
 \cmidrule(lr){3-4}
 \cmidrule(lr){5-6}
 \cmidrule(lr){7-8}
 {class} &  & {Fusion} & {DRC} & {Fusion} & {DRC} & {Fusion} & {DRC} \tabularnewline
\midrule 
{aero} & 0.57 & - & 0.50 & 0.54 & 0.49 & 0.46 & 0.51 \tabularnewline
{car} & 0.76 & - & 0.73 & 0.71 & 0.74 & 0.71 & 0.74 \tabularnewline
{chair} & 0.47 & - & 0.43 & 0.47 & 0.44 & 0.39 & 0.45
\tabularnewline
\bottomrule
\end{tabular}}

\vspace{-2mm}
\caption{Analysis of our method using mean IoU on ShapeNet.}
\vspace{-3mm}
\tablelabel{snetEval}
\end{table}

\vspace{2mm}
\noindent \textbf{Results.}
We present the results of the experiments in \tableref{snetEval} and visualize sample predictions in \figref{shapenetFig}. In general, the qualitative and quantitative results in our setting of using only a small set of multi-view observations are encouragingly close to the upper bound of using ground-truth 3D as supervision.
While our approach and the alternative way of depth fusion are comparable in the case of perfect depth information, our approach is much more robust to noisy training signal. This is because of the use of a ray potential where the noisy signal only adds a small penalty to the true shape unlike in the case of depth fusion where the noisy signal is used to compute independent unary terms (see appendix for detailed discussion).
We observe that even using only object masks leads to comparable performance to using depth but is worse when fewer views are available (\figref{snetAnalysis}) and has some systematic errors \eg the chair models cannot learn the concavities present in the seat using foreground mask information.

\vspace{2mm}
\noindent \textbf{Ablations.} When using muti-view supervision, it is informative to look at the change in performance as the number of available training views is increased. We show this result in \figref{snetAnalysis} and observe a performance gain as number of views initially increase but see the performance saturate after few views. We also note that depth observations are more informative than masks when very small number of views are used. Another aspect studied is the reconstruction performance when varying the amount of noise in depth observations. We observe that our approach is fairly robust to noise unlike the fusion approach. See appendix for further details, discussion and explanations of the trends.

\subsection{Object Reconstruction on PASCAL VOC}
We demonstrate the application of our DRC formulation on the PASCAL VOC dataset~\cite{everingham2010pascal} where previous 3D supervised single-view reconstruction methods cannot be used due to lack of ground-truth training data. However, available annotations for segmentation masks and camera pose allow application of our framework.

\vspace{2mm}
\noindent \textbf{Training Data.}
We use annotated pose (in PASCAL 3D~\cite{xiang2014beyond}) and segmentation masks (from PASCAL VOC) as training signal for object reconstruction. To augment training data, we also use the Imagenet \cite{russakovsky2015imagenet} objects from PASCAL 3D (using an off-the shelf instance segmentation method~\cite{li2015iterative} to compute foreground masks on these). These annotations effectively provide an orthographic camera $C_i$ for each training instance. Additionally, the annotated segmentation mask provides us with the observation $O_i$. We use the proposed view consistency loss on objects from the training set in PASCAL3D -- the loss measures consistency of the predicted 3D shape given training RGB image $I_i$ with the single observation-camera pair $(O_i, C_i)$. Despite only one observation per instance, the shared prediction model can learn to predict complete 3D shapes.

\vspace{2mm}
\noindent \textbf{Benchmark.}  PASCAL3D also provides annotations for (approximate) 3D shape of objects using a small set of CAD models (about 10 per category). Similar to previous approaches ~\cite{choy20163d,pamishapeTulsianiKCM15}, we use these annotations on the test set for benchmarking purposes.
Note that since the same small set of models is shared across training and test objects, using the PASCAL3D models for training is likely to bias the evaluation. 
This makes our results incomparable to those reported in~\cite{choy20163d} where a model pretrained on ShapeNet data is fine-tuned on PASCAL3D using shapes from this small set of models as ground-truth. See appendix for further discussion.

\begin{figure*}[t!]
\centering
\includegraphics[width=1.0\textwidth]{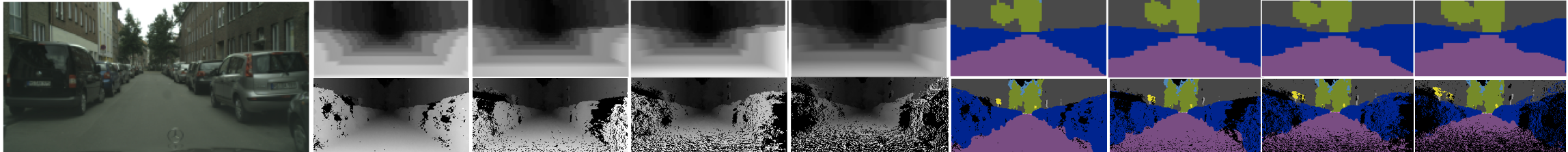}

\vspace{-2mm}
\caption{\small Sample results on Cityscapes using ego-motion sequences for learning single image 3D reconstruction. Given a single input image (left), our model predicts voxel occupancy probabilities and per-voxel semantic class distribution. We use this prediction to render, in the top row, estimated disparity and semantics for a camera moving forward by 3, 6, 9, 12 metres respectively. The bottom row renders similar output but using a 2.5D representation of ground-truth pixel-wise disparity and pixel-wise semantic labels inferred by ~\cite{dilation}.}
\vspace{-3mm}
\figlabel{csFig}
\end{figure*}

\vspace{2mm}
\noindent \textbf{Setup.}
The various baselines/variants studied are described below. Note that for all the learning based methods, we train a single category-agnostic CNN.

\vspace{1mm}
\noindent  \textit{Category-Specific Deformable Models (CSDM).} We compare to~\cite{pamishapeTulsianiKCM15} in a setting where, unlike other methods, it uses ground-truth mask, keypoints to fit deformable 3D models.

\vspace{1mm}
\noindent  \textit{ShapeNet 3D (with Realistic Rendering).} To emulate the setup used by previous approaches \eg \cite{choy20163d,Girdhar16b}, we train a CNN on rendered ShapeNet images using cross entropy loss with the ground-truth CAD model. We attempt to bridge the domain gap by using more realistic renderings via random background/lighting variations~\cite{su2015render} and initializing the convolution layers with a pretrained ResNet-18 model~\cite{he2015deep}.

\vspace{1mm}
\noindent  \textit{DRC (Pascal).} We only use the PASCAL3D instances with pose, object mask annotations to train the CNN with the proposed view consistency loss.

\vspace{1mm}
\noindent  \textit{DRC (Joint : ShapeNet 3D + Pascal).} We pre-train a model on ShapeNet 3D data as above and finetune it using PASCAL3D using our view consistency loss.

\renewcommand{\arraystretch}{1.4}
\setlength{\tabcolsep}{10pt}
\begin{table}
\centering
\small
\begin{tabular}{lcccc}
\toprule
{Method} & {aero} & {car} & {chair} & {mean}
 \\
 \cmidrule(lr){1-1}
 \cmidrule(lr){2-4}
 \cmidrule(lr){5-5}
{CSDM} & 0.40 & 0.60 & 0.29 & 0.43 \tabularnewline
{DRC (PASCAL)} & 0.42 & 0.67 & 0.25 & 0.44 \tabularnewline
{Shapenet 3D} & 0.53 & 0.67 & 0.33 & 0.51 \tabularnewline
{DRC (Joint)} & \textbf{0.55} & \textbf{0.72} & \textbf{0.34} & \textbf{0.54}
\tabularnewline
\bottomrule
\end{tabular}

\vspace{-2mm}
\caption{Mean IoU on PASCAL VOC.}
\vspace{-3mm}
\tablelabel{pascalEval}
\end{table}

\vspace{2mm}
\noindent \textbf{Results.}
We present the comparisons of our approach to the baselines in \tableref{pascalEval} and visualize sample predictions in \figref{pascalFig}. We observe that our model when trained using only PASCAL3D data, while being category agnostic and not using ground-truth annotations for testing, performs comparably to \cite{pamishapeTulsianiKCM15} which also uses similar training data. We observe that using the PASCAL data via the view consistency loss in addition to the ShapeNet 3D training data allows us to improve across categories as using real images for training removes some error modes that the CNN trained on synthetic data exhibits on real images. Note that the learning signals used in this setup were only approximate -- the annotated pose, segmentation masks computed by \cite{li2015iterative} are not perfect and our method results in improvements despite these.

\subsection{3D Scene Reconstruction from Ego-motion}
The problem of scene reconstruction is an extremely challenging one. While previous approaches, using direct~\cite{eigen2015predicting}, multi-view~\cite{garg2016unsupervised,godard2016unsupervised} or even no supervision~\cite{Fouhey15} predict detailed 2.5D representations (pixelwise depth and/or surface normals), the task of single image 3D prediction has been largely unexplored for scenes. A prominent reason for this is the lack of supervisory data. Even though obtaining full 3D supervision might be difficult, obtaining multi-view observations may be more feasible. We present some preliminary explorations and apply our framework to learn single image 3D reconstruction for scenes by using driving sequences as supervision.

We use the cityscapes dataset~\cite{cityscapes} which has numerous 30-frame driving sequences with associated disparity images, ego-motion information and semantic labels\footnote{while only sparse frames are annotated, we use a semantic segmentation system~\cite{dilation} trained on these to obtain labels for other frames}. We train a CNN to predict, from a single scene image, occupancies and per-voxel semantic labels for a coarse  voxel grid. We minimize the consistency loss function corresponding to the event cost in \eqref{psisem}. To account for the large scale of scenes, our voxel grid does not have uniform cells, instead the size of the cells grows as we move away from the camera. See appendix for details, CNN architecture \etc.

We show qualitative results in \figref{csFig} and compare the coarse 3D representation inferred by our method with a detailed 2.5D representation by rendering inferred disparity and semantic segmentation images under simulated forward motion. The 3D representation, while coarse, is able to capture structure not visible in the original image (\eg cars occluding other cars). While this is an encouraging result that demonstrates the possibility of going beyond 2.5D for scenes, there are several challenges that remain \eg the pedestrians/moving cars violate the implicit static scene assumption, the scope of 3D data captured from the multiple views is limited in context of the whole scene and finally, one may never get observations for some aspects \eg multi-view supervision cannot inform us that there is road below the cars parked on the side.

\begin{figure}[t!]
\hspace{-7mm}
\includegraphics[width=0.53\textwidth]{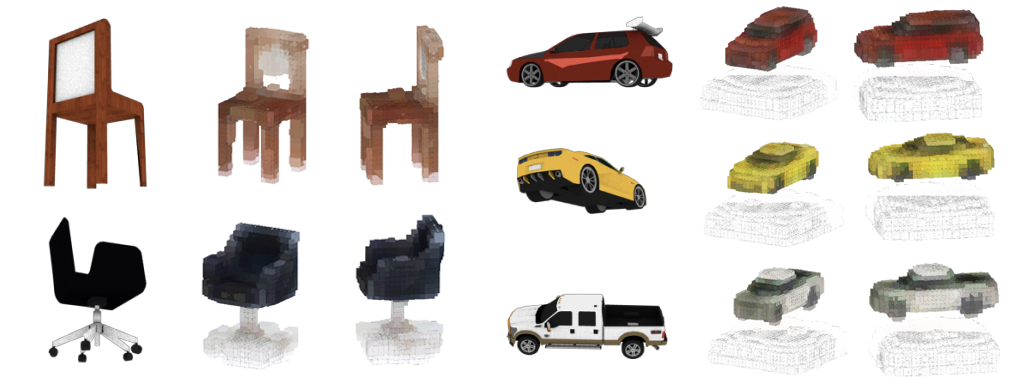}

\vspace{-2mm}
\caption{Sample results on ShapeNet dataset using multiple RGB images as supervision for training. We show the input image (left) and the visualize 3D shape predicted using our learned model from two novel views. Best viewed in color.}
\vspace{-3mm}
\figlabel{snetColorFig}
\end{figure}

\subsection{Object Reconstruction from RGB Supervision}
We study the setting where only 2D color images of ShapeNet models are available as supervisory signal. In this scenario, our CNN predicts a per-voxel occupancy as well as a color value. We use the generalized event cost function from \eqref{psicolor}~to define the training loss. Some qualitative results are shown in \figref{snetColorFig}. We see the learned model can infer the correct shape as well as color, including the concavities in chairs, shading for hidden parts \etc. See appendix for more details and discussion on error modes \eg artifacts below cars.

\vspace{-1mm}
\section{Discussion}
\vspace{-1mm}
We have presented a differentiable formulation for consistency between a 3D shape and a 2D observation and demonstrated its applications for learning single-view reconstruction in various scenarios. These are, however, only the initial steps and a number of challenges are yet to be addressed.
Our formulation is applicable to voxel-occupancy based representations and an interesting direction is to extend these ideas to alternate representations which allow finer predictions \eg~\cite{hane2017hierarchical,riegler2017octnetfusion,tatarchenko2017octree}.
Additionally, we assume a known camera transformation across views. While this is a realistic assumption from the perspective of agents, relaxing this might further allow learning from web-scale data.
Finally, while our approach allows us to bypass the availability of ground-truth 3D information for training, a benchmark dataset is still required for evaluation which may be challenging for scenarios like scene reconstruction.

\vspace{2mm}
\noindent \textbf{Acknowledgements.}
{We thank the anonymous reviewers for helpful comments. This work was supported in part by Intel/NSF VEC award IIS-1539099, NSF Award IIS-1212798, and the Berkeley Fellowship to ST. We gratefully acknowledge NVIDIA corporation for the donation of Tesla GPUs used for this research.}

{\small
\bibliographystyle{ieee}
\bibliography{cvpr17drc}
}

\clearpage
\setcounter{page}{1}
\title{Appendix : Multi-view Supervision for Single-view Reconstruction\\ via Differentiable Ray Consistency
\vspace{-0.5cm}}
\author{Shubham Tulsiani, Tinghui Zhou, Alexei A. Efros, Jitendra Malik  \\
University of California, Berkeley\\
{\tt\small\{shubhtuls,tinghuiz,efros,malik\}@eecs.berkeley.edu}
}
\setlength{\droptitle}{-2cm}
\maketitle
\section*{A1. Gradient Derivations}
We re-iterate the equations for event probabilities and the  ray consistency loss as defined in the main text.
\begin{gather}
\eqlabel{pzr_rep}
p(z_r = i) = \begin{dcases}
    (1 - x^r_i) \prod_{j=1}^{i-1} x^r_j, & \text{if } i \leq N_r\\
     \quad \prod_{j=1}^{N_r} x^r_j,  & \text{if } i = N_r+1
\end{dcases} \\
\eqlabel{Lrdef2_rep}
L_r(x) = \sum_{i=1}^{N_r+1} \psi_r(i)~p(z_r = i)
\end{gather}

Expanding \eqref{Lrdef2_rep} using \eqref{pzr_rep}, we can get --
\begin{align*}
L_r(x) = \sum_{i=1}^{N_r} \psi_r(i)~(1 - x^r_i) \prod_{j=1}^{i-1} x^r_j ~+~\psi_r(N_r + 1)~ \prod_{j=1}^{N_r} x^r_j \\
 = \sum_{i=1}^{N_r+1} \psi_r(i) \prod_{j=1}^{i-1} x^r_j ~-~\sum_{i=1}^{N_r} \psi_r(i) \prod_{j=1}^{i} x^r_j \\
 = \psi_r(1) +  \sum_{i=1}^{N_r} \psi_r(i+1) \prod_{j=1}^{i} x^r_j ~-~\sum_{i=1}^{N_r} \psi_r(i) \prod_{j=1}^{i} x^r_j
\end{align*}

\noindent Simplifying this, we finally obtain --
\begin{gather}
\eqlabel{simpleL}
L_r(x) =  \psi_r(1) + \sum_{i=1}^{N_r} ~ (\psi_r(i+1) - \psi_r(i)) \prod_{j=1}^{i} x^r_j
\end{gather}

\noindent We can now compute the derivatives of the ray consistency loss w.r.t. the predictions $x$ --
\begin{align*}
\frac{\partial~L_r(x)}{\partial~x^r_k} = \sum_{i=1}^{N_r} ~ (\psi_r(i+1) - \psi_r(i)) ~ \frac{\partial~\prod_{j=1}^{i} x^r_j}{\partial~x^r_k}   \\
 = \sum_{i=k}^{N_r} ~ (\psi_r(i+1) - \psi_r(i)) \prod_{1\leq j \leq i, j \neq k} x^r_j
\end{align*}

\section*{A2. Additional Discussion}
\subsection*{A2.1. Formulation}
\vspace{2mm}
\noindent \textbf{Relation with Reprojection Error for Mask Supervision.}
In the scenario with foreground mask supervision, we had defined the corresponding cost term as in \eqref{psiMaskRep}.
\begin{gather}
\eqlabel{psiMaskRep}
\psi^{mask}_r(i) = 
\begin{dcases}
    s_r, & \text{if } i \leq N_r\\
    1-s_r,  & \text{if } i = N_r+1
\end{dcases}
\end{gather}
Here, $s_r \in \{0,1\}$ denotes the known information regarding each ray (or image pixel) - $s_r=0$ implies the ray $r$ intersects the object \ie corresponds to an image pixel with foreground label, $s_r=1$ indicates a pixel with background label. We observe that using this definition of event cost function, we can further simplify the loss defined in \eqref{simpleL}.
\begin{align*}
L^{mask}_r(x) & =  \psi_r(1) + \sum_{i=1}^{N_r} ~ (\psi_r(i+1) - \psi_r(i)) \prod_{j=1}^{i} x^r_j \\
\\ & = s_r + ~(~\sum_{i=1}^{N_r-1} ~ (s_r - s_r) \prod_{j=1}^{i} x^r_j ~)~ + (1 - 2s_r) \prod_{i=1}^{N_r} x^r_i
\\ & = s_r + (1 - 2s_r) \prod_{i=1}^{N_r} x^r_i
 \\ &= | \prod_{i=1}^{N_r} x^r_i  - s_r|,~~~\text{if}~s_r \in \{0,1\}
\end{align*}
Therefore, we see that in the case of foreground mask observations, we minimize the discrepancy between the observation for the ray \ie $s_r$ and the reprojected occupancies ($\prod_{i=1}^{N_r} x^r_i$). This is similar to concurrent approaches designed to specifically use mask supervision for object reconstruction where a learned~\cite{rezende2016unsupervised} or fixed~\cite{yan2016perspective} reprojection function is used. In particular, the reprojection function ($\min_{i=1}^{N_r} x^r_i$) used by Yan \etal~\cite{yan2016perspective} can be thought of as an approximation of the one derived using our formulation.

\vspace{2mm}
\noindent \textbf{Event Costs for $z_r = N_r + 1$.}
We noted that to instantiate the event cost functions, we need to define the induced observations for the event of the ray escaping. While this is simple for some scenarios \eg mask supervision, we need to define the induced observations ($d^r_{N_r+1}, p^r_{N_r+1}$ \etc.) in other cases more carefully. When using depth observations for object reconstruction, we defined $d^r_{N_r+1}$ to be a fixed large value (10m as opposed to $\infty$) as we were penalizing absolute difference. Similarly, we used a uniform distribution as $p^r_{N_r+1}$ when using semantics observation so that the log-likelihood error is bounded. When using RBG supervision, we leveraged the knowledge that renderings had a white background to define $p^r_{N_r+1}$ -- one could argue this is akin to implicitly using mask supervision but note that a) some models can also have white texture, b) our RGB-supervised model learned concavities which pure mask supervision could not, and c) we recover additional information (color) per voxel.

We wish to highlight here the takeaway that to completely define an event cost function, one must specify how the event of ray escaping is handled. While here we used a fixed induced observation corresponding to ray escaping events, this could equivalently be predicted in addition to the 3D shape \eg predicting an environment map which allows us to lookup the color for an escaping ray conditioned on the direction can allow modelling the 3D shape and background with multi-view RGB observations.

\vspace{2mm}
\noindent \textbf{Application Scope.} We show four kinds of observations that can be leveraged via our framework -- masks, depth, color and semantics. An obvious question is whether we can characterize scenarios the where the presented formulation may be applicable. As the main text notes, one fundamental assumption is the availability of relative pose between the prediction frame and observations. An aspect that is harder to formalize is what kinds of `observations' can be used for supervision. We note that the only requirement is that an event cost function $\psi_r(i, [p_i^r])$ be defined such that it is differentiable w.r.t $p_i^r$. This implies that the known ray observation $o_r$ should be `explainable' given the ray (\ie origin and direction), the location of the $i^{th}$ voxel on its path, and some (view, ray agnostic) auxiliary prediction for the voxel.


\subsection*{A2.2. Experimental Setup}
\vspace{2mm}
\noindent \textbf{Biases in `Ground-truth' Voxelization.} We would like to mention a subtle practical aspect of benchmarking multi-view supervised approaches. The `ground-truth' voxelization used for evaluation is binary. To obtain this binary label, typical voxelization methods (including the one we used) consider a grid cell occupied if there is any part of surface inside it. This is notably different from the information that view supervised approaches would obtain -- a partially occupied cell would receive evidences for being free as well as for being occupied, as some rays would pass through it and some would not. Therefore, the binary `ground-truth' used for evaluation is biased to be over-inflated. It should be noted that the 3D supervised approaches have this bias in the supervision (as they use voxelizations computed similarly as ground-truth) whereas multi-view supervised approaches do not -- we think this, at least partially, explains the gap between our multi-view supervised approach when using large number of views and 3D supervision. Note that previous mask-supervised approaches \eg Yan \etal~\cite{yan2016perspective} did not suffer from this as they do not use actual rendered images for supervision but instead just use projections of the (biased) voxelizations as ground-truth mask `views'. While in this work we use an IoU based metric, a metric that can handle a soft ground-truth might be more appropriate.

\vspace{2mm}
\noindent \textbf{On Benchmarking Protocols for PASCAL 3D Dataset.} Another subtle aspect we would like to highlight is that the `ground-truth' 3D annotations on PASCAL3D dataset are obtained by choosing the appropriate model from a small set of CAD models. In particular, these models are shared for annotating train and test instances. Therefore, a benchmarking protocol where a learning based method has access to these models for training can bias the performance. To further highlight this point, we conducted a simple experiment. PASCAL3D has 65 CAD models across all categories (with 8, 10, 10 models respectively for aeroplanes, cars and chairs). Using the annotations on the training set, we train a classifier (by finetuning a pretrained ResNet-18 model) to choose the annotated model corresponding to an object given a cropped bounding box image. We then `reconstruct' a test set object by  retrieving the predicted model. We obtain IoUs of $(0.74, 0.76, 0.46)$ respectively for aeroplane, car and chair categories -- significantly higher than our approach. Across the 10 object categories, the mean IoU via retrieval is $0.72$.

The point we would like to emphasize is that using the small set of CAD models for both, training and testing will reward a learning system for biasing itself towards these small set of models and is therefore not a recommended benchmarking protocol. Note that since we train using masks and pose for supervision, we do not leverage these models for training and only use them for evaluation.

\section*{A3. CNN Architectures, Experiment Details and Result Interpretation}
For all the object-reconstruction experiments, similar to previous learning based methods~\cite{choy20163d,Girdhar16b}, our CNN is trained to output reconstructions in a canonical frame (where object is front-facing and upright) -- the known camera associated with each observation image is assumed w.r.t. this canonical frame. For the scene reconstruction experiment, we output the reconstructions w.r.t. the frame associated with the camera corresponding to the input image. We therefore assume known relative transformations between the input image frame and the used observations.

As mentioned in the main text, to efficiently implement the view consistency loss between the prediction and an observation image, we randomly sample rays. We always sample total 3000 rays (chosen to keep iteration time under 1 sec) for each instance in the mini-batch \ie if we have $k$ observation images, we sample $\frac{3000}{k}$ rays per observation. Additionally, for efficiency in the scene reconstruction experiment, instead of using all 30 observations in the sequence, we randomly sample 3 observations per iteration (and sample 1000 rays per observation). For ShapeNet experiments, we use all available observations (typically 5) -- for the ablation where we vary number the number of observations available, we use all $k$ observations in each minibatch. Finally, we observed that a large fraction of rays in the object reconstruction experiments (ShapeNet and PASCAL VOC) corresponded to background pixels. To counter this, we weight the loss for the foreground rays higher by a factor of 5 (hyperparameter chosen using chair reconstruction evaluation on the validation set using Mask observations on ShapeNet) -- an alternate would be to simply sample more foreground rays.

\begin{figure*}[t!]
\centering
\includegraphics[width=1.0\textwidth]{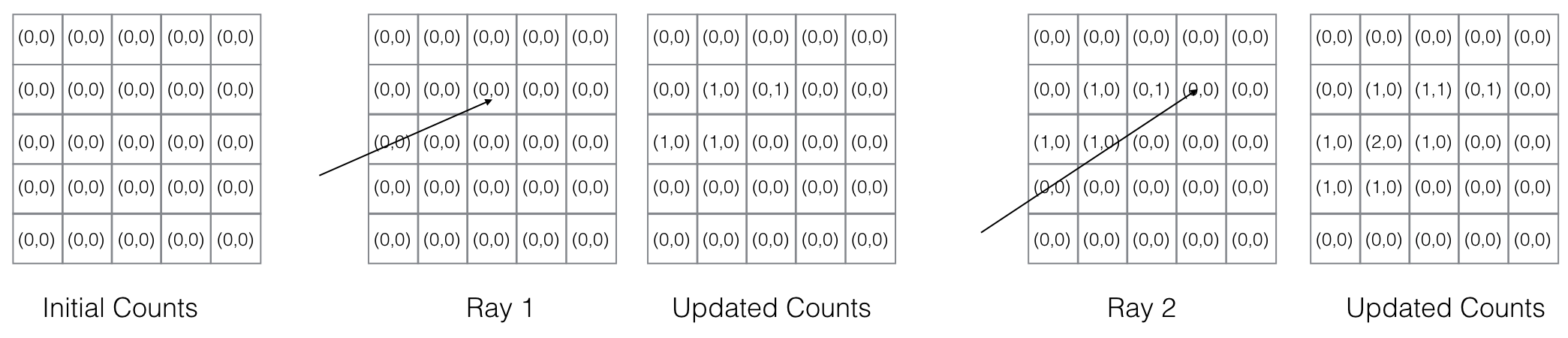}
\caption{We show the process of precomputing a pseudo-groundtruth 3D model using available depth images (denoted as the `Depth Fusion' baseline in main text). We maintain (empty,occupied) count for each voxel which respectively indicate the number of rays that pass through the voxel or terminate in it. Given the observed depth images, we can compute the starting and termination points of all rays. We show an example of passing two rays and updating the maintained counts. The final counts are used to determine a soft occupancy value for each voxel which serves as the prediction target -- we ignore the voxels where both counts are 0.}
\figlabel{depthFusion}
\end{figure*}

In all the experiments, the basic network architecture is of an encoder-decoder form. The encoder takes as input an RGB image and uses 2D convolutions and fully-connected layers to encode it. The decoder performs 3D up-convolutions to finally predict the voxel occupancies (and optional per-voxel predictions). To succinctly describe the architectures, we denote by $C(n)$ a convolution layer with $3~X~3$ kernel and $n$ output channels, followed by a $ReLU$ non-linearity and spatial pooling such that the spatial dimensions reduce by a factor of two. We denote by $R()$ a reshape layer that reshapes the input according to the size specified when instantiating it. Similarly by $UC(n)$, we denote a 3D up-convolution followed by $ReLU$, where the output's spatial size is doubled. Unless otherwise specified, we initialize all weights randomly and use ADAM for training the networks.

\subsection*{A3.1. ShapeNet (Mask/Depth Supervision)}

\noindent \textbf{Architecture.}
The CNN architecture used here is extremely simple and has an encoder $Img(3,64,64)-C(8)-C(16)-C(32)-C(64)-C(128)-FC(100)-FC(100)-FC(128)-R(16,2,2,2)$. The decoder is of the form $UC(8)-UC(4)-UC(2)-UC(1)-\text{Sigmoid}()$. The final voxel grid  has 32 cells in each dimension. Note that since the aim here is to analyze the approach in a simple setting, we train a separate CNN per category, unlike the PASCAL VOC experiments where we follow previous work to train a category-agnostic CNN.

\vspace{2mm}
\noindent \textbf{Rendered Data.}
To obtain the RGB/Depth/Mask images which are used as input for the CNN or as `observations' for training using the view consistency loss, we render the ShapeNet models using Blender. For each CAD model, we render images from random viewpoints obtained via uniformly sampling azimuth and elevation from $[0,360)$ and $[-20,30]$ degrees respectively. We also use random lighting variations to render the RGB images. To obtain noisy depth observations, we add uniform and independent per-pixel noise to the rendered depth images. The settings reported in the experiment table correspond to a maximum noise of 20cm but we also separately study other amounts of maximum noise. To interpret the amount of noise, note that the models in ShapeNet are normalized to be in a 1m cube.

\vspace{2mm}
\noindent \textbf{Analysis.}
The experiments and ablations revealed some interesting trends. Our DRC approach is  robust to observation noise and using about 4-5 observations saturates performance. While this performance is still below using ground-truth supervision, we conjecture this may be to due to the inflation bias in the `ground-truth'. Also, using mask supervision and depth supervision are both equally informative when considering a large (more than 5) number of views though depth images are more informative when using 1-2 observations.

\subsection*{A3.2. PASCAL VOC Reconstruction}
The decoder architecture used in these experiments is similar to the ShapeNet architecture described above. However, in the encoder, we use the convolution layers from ResNet-18 (initialized with copied weights from a pre-trained Imagenet classification network).

\subsection*{A3.3. Cityscapes Reconstruction}
\noindent \textbf{Architecture.}
Since the images in this dataset typically have a higher field of view in the $x$ direction than the $y$ direction, we use the architecture exactly same as the PASCAL VOC experiment (\ie with ResNet layers) with the modification that the last two layers of the encoder are $FC(256)-R(16,4,2,2)$. This allows the final voxel prediction to have more cells (64) in the $x$ dimension. In addition to voxel occupancies, the last layer also outputs a per-voxel semantic class distribution (and applies a Softmax instead of Sigmoid to these).

\vspace{2mm}
\noindent \textbf{Grid Parametrization.}
Since outdoor scenes have a much larger spatial extent, we use a non-uniform voxel grid in this experiment. The cell sizes increase as the $z$ coordinate increases. This has the effect of modeling the nearby structures with more resolution but the far-away structures are captured only coarsely. The voxel-coordinate system lies in the range $x \in [0,64], y \in [0,32], z \in [0,32]$ with the cell centres located at integer locations offset by $\frac{1}{2}$. The correspondence of a coordinate $(x,y,z)$ of this system to the real world is given by $p = \alpha_1e^{\alpha_2 z} (f(x-32),f(y-16),1)$. Note that the volumetric bins are uniform when projected into an image. Here $\alpha_1, \alpha_2$ are defined s.t. the minimum and maximum $z$ coordinates are 0.5m are 1000m. The parameter $f$ is defined to give a horizontal field of view of 50 degrees.

\subsection*{A3.4. RGB-Supervised Reconstruction}
\noindent \textbf{Architecture.}
The CNN architecture here is the same as the one used for 3D prediction on ShapeNet with depth/mask supervision with the change that the last $UC$ layer predicts 4 channels instead of 1 where the last three correspond to per-voxel color predictions.

\vspace{2mm}
\noindent \textbf{Analysis.}
We observed that our learned model using multi-view color image observations as supervision could recover, from a single input image, the shape and texture of the full 3D shape including details like concavities in chairs, texture on unseen surfaces \etc. An interesting error mode however, was a white cloud-like structure predicted below cars. This was predicted because, due to limited elevation variation in our view sampling, all the rays that pass through that region actually correspond to the background and so have a white color associated with them. This observation can be `explained' via two solutions - a) to predict empty space and allow the rays to escape, and b) to predict occupied space with white texture. Our learned CNN chooses the latter way of explaining the multi-view observations.

\section*{A4. Depth Fusion}
We show in \figref{depthFusion} the process of computing the pseudo-ground truth shape used for our baseline. Note that this is applicable only for depth images but not object mask observations since the termination points of the rays are unknown in case of foreground masks. This process is analogous to extracting and averaging unary terms for voxel occupancies for each ray -- the voxels that rays pass through are assumed to be empty and the ones where they terminate occupied.

This process does not model any noise in observations unlike ours, which has a higher-order cost term. To clarify this, consider a ray for which we have a (noisy) depth observation. Let $v1, v2, v3$ be 3 voxels (in order) in its path and let the observed depth imply termination in $v3$. Suppose that the `true' model has $v2$ occupied and not $v3$. Under our cost term, the `true' shape would incur only a small cost for the noisy observation but under if we only have unary terms as in the fusion approach, this would incur a high cost. 

\end{document}